\documentclass[12pt]{article}

\usepackage{amsthm}
\usepackage{graphicx}
\usepackage{algorithm}
\usepackage{algorithmicx}
\usepackage{algpseudocode}

\usepackage{times}

\newenvironment{sciabstract}{%
\begin{quote} \bf}
{\end{quote}}

\newcounter{lastnote}

\title{Using Gaussian Measures for Efficient Constraint Based Clustering}

\author
{Chandrima Sarkar,$^{1\ast}$ Atanu Roy,$^{1}$ \\
\normalsize{sarkar@cs.umn.edu, atanu@cs.umn.edu}\\
\normalsize{$^{1}$ Department of Computer Science and Engineering, University of Minnesota, Twin Cities}\\
\normalsize{$^\ast$To whom correspondence should be addressed; E-mail:  sarkar@cs.umn.edu}
}

\date{}

\begin{document} 


\baselineskip24pt


\maketitle


\begin{sciabstract}
  In this paper we present a novel iterative multiphase clustering technique
for efficiently clustering high dimensional data points. For this
purpose we implement clustering feature (CF) tree on a real data set
and a Gaussian density distribution constraint on the resultant CF
tree. The post processing by the application of Gaussian density distribution
function on the micro-clusters leads to refinement of the previously
formed clusters thus improving their quality. This algorithm also
succeeds in overcoming the inherent drawbacks of conventional hierarchical
methods of clustering like inability to undo the change made to the
dendogram of the data points. Moreover, the constraint measure applied
in the algorithm makes this clustering technique suitable for need
driven data analysis. We provide veracity of our claim by evaluating
our algorithm with other similar clustering algorithms. 
\end{sciabstract}


\section*{Introduction}

In this paper we examine clustering, which is an important category of data mining problem. Data mining is the process of finding hidden patterns and trends in databases and using that information to do a variety of tasks such as finding association rules, clustering heterogeneous groups of information and build predictive models \cite{koh2011data}. It can also be considered as an important tool for data segmentation, selection, exploration and building models using the vast data stores to discover previously unknown patterns in various domains such as healthcare \cite{milley2000healthcare} \cite{sarkar2013correlation}, \cite{sarkar2013impact}, \cite{sarkar2012improved} \cite{sarkar2014robust},  social media analysis \cite{sarkar2014feature},\cite{DBLP:journals/corr/SinghalRS14}, \cite{DBLP:conf/asunam/AhmadKRWSC13}, \cite{DBLP:conf/asunam/RoyBS13}, \cite{DBLP:conf/nsw/LeungDHAPHWSMRS13}, \cite{DBLP:conf/socialcom/RoyASKS12}, \cite{DBLP:conf/icdm/RoySA10},  finances and various other domains \cite{sarkar2014predictive}. From the past few decades, data mining has been used extensively in various areas of decision making and decision analysis by financial institutions, for credit scoring and fraud detection; marketers, for direct marketing and cross-selling or up-selling; retailers, for market segmentation and store layout; and manufacturers, for quality control and maintenance scheduling. Little has been explored in the healthcare domain using data mining. Clustering can be
defined as division of data into groups of similar objects without
any prior knowledge. Though representation of the data objects by
clusters can lead to a loss of certain finer details, simplification
can be achieved at the same time. From a machine learning perspective
clusters correspond to hidden patterns. The search for clusters is
an example of unsupervised learning. From a practical perspective
clustering plays an important role in data mining applications such
as scientific data exploration, information retrieval and text mining,
spatial database applications, web analysis, CRM, marketing, medical
diagnostics, computational biology and many others.

Data clustering algorithms can be further sub divided into a number
of categories. In this paper we concentrate on hierarchical and constraint
based clustering. With hierarchical algorithms, successive clusters
are found using previously established clusters, whereas constraint
based clustering algorithms use some pre-defined constraints to create
the clusters. Hierarchical algorithms can be agglomerate (bottom-up)
or divisive (top-down). Agglomerate algorithms begin with each element
as a separate cluster and merge them in successively larger clusters.
The divisive algorithms begin with the whole set and proceed to divide
it successively into smaller clusters. A hierarchical algorithm in
general yields a deprogram representing the nested grouping of patterns
and similarity levels at which these groupings change. But there are
some inherent problems associated with greedy hierarchical algorithmic
approaches (AGNES, DIANA) \cite{handatamining2nded} like vagueness
of termination criteria of the algorithms and the inability to revisit
once constructed clusters for the purpose of their improvement. To
overcome these defects, an efficient technique called the multiphase
clustering technique \cite{DBLP:conf/sigmod/ZhangRL96} can be used.
This technique involves two primary phases \textendash{} phase 1,
where the database is scanned to built a temporary in-memory tree
in order to preserve the inherent clustering structure of the data.
Phase 2 selects a clustering algorithm and apply it to the leaf nodes
of the tree built in the previous phase for removing sparse clusters
as outliers and dense clusters into larger ones. Our approach uses
this technique to build an efficient hierarchical constraint based
clustering method.

Constraint based clustering is a clustering technique which deals
with specific application requirements. It falls in a category in
which we can divide the clustering methods into either data-driven
or need-driven \cite{Banerjee_scalableclustering}. The data-driven
clustering methods intend to discover the true structure of the underlying
data by grouping similar objects together while the need-driven clustering
methods group objects based on not only similarity but also needs
imposed by a particular application. Thus, the clusters generated
by need-driven clustering are usually more useful and actionable to
meet certain application requirements. For different application needs,
various balancing constraints can be designed to restrict the generated
clusters and make them actionable. Particularly, we are interested
in such constraint types for phase 2 of our hierarchical clustering.
In this paper we will examine Gaussian distribution measures as the
density based probabilistic measures which will decide the threshold
for split of the leaf node clusters formed by the CF tree in phase
1. By imposing a threshold of minimum Gaussian distribution value
on the clusters, our model will search for clusters which are balanced
in terms of probabilistic density distribution measure. The main motivation
of our paper is to develop an approach which efficiently implements
the multiphase clustering technique using Gaussian distribution measure
as the constraint for creation of efficient clusters. In phase 1 first
we aim at implementing a data structure similar to the CF-tree \cite{DBLP:conf/sigmod/ZhangRL96}.
In the next phase we apply our Gaussian constraint for efficient clustering. 

The main contribution of our paper are as follows:
\begin{enumerate}
\item We propose the multiphase clustering technique using a data structure
like CF-tree and a Gaussian distribution measure as the constraint
in phase 2.
\item We evaluate our algorithm using supervised clustering evaluation measure
and compare it with some existing clustering techniques.
\item We also evaluate our algorithm over real data set for demonstrating
the quality of the generated cluster and efficiency of the algorithm.
\end{enumerate}
The remainder of the paper is organized as follows. Section II provides
a list of the related works. Section III discusses in detail the proposed
approach. Section IV is devoted to experimental results. We conclude
our research and put forth our future works in Section V.

\section{Literature Review}

Clustering has been an active research area in the fields of data
mining and computational statistics. Among all of the clustering methods,
over the past few years, constrained (semi-supervised) based clustering
methods have become very popular. Clustering with constraints is an
important recent development in the clustering literature. The addition
of constraints allows users to incorporate domain expertise into the
clustering process by explicitly specifying what are desirable properties
in a clustering solution. Constraint based clustering takes advantage
of the fact that some labels (cluster membership) are known in advance,
and therefore the induced constraints are more explicit. This is particularly
useful for applications in domains where considerable domain expertise
already exists. It has been illustrated by several researchers that
constraints can improve the results of a variety of clustering algorithms
\cite{Zhu:2010:DCS:1857275.1857774}. However, there can be a large
variation in this improvement, even for a fixed number of constraints
for a given data set. In this respect, \cite{Wagstaff06whenis} has
shown that inconsistency and incoherence are two important properties
of constraint based measures. These measures are strongly anti-correlated
with clustering algorithm performance. Since they can be computed
prior to clustering, these measures can aid in deciding which constraints
to use in practice. 

In real world applications such as image coding clustering, spatial
clustering, geoinformatics \cite{DBLP:conf/dgo/PatilBAJ07}, and document
clustering \cite{Laguia:2008:LDC:1411852.1412098}, some background
information is usually obtained dealing with the data objects\textquoteright{}
relationships or the approximate size of each group before conducting
clustering. This information is very helpful in clustering the data
\cite{Zhu:2010:DCS:1857275.1857774}. However, traditional clustering
algorithms do not provide effective mechanisms to make use of this
information. Previous research has looked at using instance-level
background information, such as must-link and cannot-link constraints
\cite{Basu04activesemi-supervision,Wagstaff01constrainedk-means}.
If two objects are known to be in the same group, we say that they
are must-linked, and if they are known to be in different groups,
we say that they are cannot-linked. Wagstaff\textit{ et al.} \cite{Wagstaff01constrainedk-means}
incorporated this type of background information to k-means algorithm
by ensuring that during the clustering process, at each iteration,
each constraint is satisfied. In cannot-link constraints, the two
points cannot be in the same cluster. Basu \textit{et al.} \cite{Basu04activesemi-supervision}
also considered must-link and cannot-link constrains to learn an underlying
metric between points while clustering. 

There are also works on different types of knowledge hints used in
fuzzy clustering, including partial supervision where some data points
have been labeled \cite{DBLP:journals/tsmc/PedryczW97} and domain
knowledge represented in the form of a collection of viewpoints (e.g.,
externally introduced prototypes/ representatives by users) \cite{Zhong_scalablebalanced}.
Another type of constraints which has been dealt with in the recent
past is balancing constraints in which clusters are of approximately
the same size or importance \cite{Banerjee_scalableclustering,DBLP:conf/icml/HellerG05}.
Constraint driven clustering has been also investigated by \cite{DBLP:conf/kdd/GeEJD07},
where the author proposes an algorithm based on \cite{DBLP:conf/sigmod/ZhangRL96}
with respect to two following constraints \textendash{} minimum significance
constraints and minimum variance constraints. All the above mentioned
researches involve specifying the number of constraint, for the formation
of clusters. However, little work has been reported on constraints
which make use of the Gaussian density distribution function to ascertain
the probability of a data point to belong to a particular cluster. 

In this paper, we aim at developing a multiphase iterative clustering
algorithm using a data structure based on the CF-tree proposed by
Zhang \textit{et al.} \cite{DBLP:conf/sigmod/ZhangRL96} and a constraint
based Gaussian density distribution function. Our algorithm will generate
clusters with respect to our proposed multivariate Gaussian distribution
constraint. Multivariate Gaussian distribution constraint is based
on the normal density distribution of the data which can be used to
ascertain the probability of that data point belonging to that cluster.
The main goal of the constraint would be to check if it is possible
to dynamical split the clusters formed in phase 1 of our algorithm
based on an assumed threshold value so that it is advantageous for
a more accurate clustering.

\section{Proposed Approach}

\subsection{Preliminaries}

Finding useful patterns in large databases without any prior knowledge
has attracted a lot of research over time \cite{DBLP:conf/sigmod/ZhangRL96,DBLP:conf/kdd/GeEJD07}.
One of widely studied problems is the identification of the clusters.
There are a lot of methods which can be used for data clustering namely
partitioning methods, grid based methods, hierarchical methods, density
based methods, model based methods, and constraint based models \cite{handatamining2nded}.
The ones that we are primarily interested in this research are the
hierarchical and constraint based models.

\subsubsection{Centroid, Radius and Diameter}

Given \textit{n} \textit{d}-dimensional data points, the centroid
of the cluster is defined as the average of all points in the cluster
and can be represented as
\begin{equation}
x_{0}=\frac{\sum_{i=0}^{n}x_{i}}{n}\label{eq:centroid}
\end{equation}

The radius of a cluster is defined as the average distance between
the centroid and the member values of the cluster whereas the diameter
is defined as the average distance between the individual members
in the cluster

\begin{equation}
R=\sqrt{\frac{\sum_{i=1}^{n}(x_{i}-x_{0})^{2}}{n}}\label{eq:radius}
\end{equation}
\begin{equation}
D=\sqrt{\frac{\sum_{i=1}^{n}\sum_{j=1}^{n}(x_{i}-x_{j})^{2}}{n(n-1)}}\label{eq:Diameter}
\end{equation}
Although equations \ref{eq:centroid},\ref{eq:radius} ,\ref{eq:Diameter}
are defined in \cite{DBLP:conf/sigmod/ZhangRL96,handatamining2nded},
but we redefine the concepts in this paper since they form an integral
part of the paper.

\subsubsection{CF-tree\label{sub:CF-Tree}}

In \cite{DBLP:conf/sigmod/ZhangRL96}, the authors use an innovative
data structure called the CF-tree (clustering feature) to summarize
information about the data points. Clustering feature is a 3-dimensional
data structure \cite{handatamining2nded} containing the number of
data points, their linear sum $(LS)$ and their squared sum $(SS)$
which can also be represented as $\sum_{i=1}^{n}x_{i}$ and $\sum_{i=1}^{n}x_{i}^{2}$
respectively. 
\begin{equation}
CF=\langle n,\emph{\emph{LS,SS\ensuremath{\rangle}}}\label{eq:CFeq}
\end{equation}

\subsubsection{Gaussian Distribution}

In the theory of probability, Gaussian or normal distribution is defined
as a continuous probability distribution of real valued random variables
which tend to cluster around the centroid of a particular class.

Multivariate (k-dimensional) Gaussian distribution is a generalization
of univariate normal distribution. It is generally written using the
following notation: $X\sim N_{k}(\mu,\Sigma)$. $X$ refers to the
$k$-dimensional vector $(X_{1},X_{2},\cdots,X_{k})$. $\mu$ refers
to the $k$-dimensional mean vector and $\sum$ refers to the $k$
x $k$ co-variance matrix. Thus the normal distribution of a matrix
where each row is an observation and each column is an attribute can
be written as:

\begin{equation}
f_{X}(x)=\frac{1}{(2\pi)^{k/2}|\Sigma|^{1/2}}exp(-\frac{1}{2}(x-\mu)'\Sigma^{-1}(x-\mu))\label{eq:mvnpdf}
\end{equation}

\subsection{Overview}

We start our algorithm with an input data set. Since our algorithm
is best suited for continuous valued variables, we use attribute selection
measure to weed out the non-continuous attributes. The resultant data
set is used as the input for the first phase of our algorithm. In
this phase we create a CF-tree originally proposed by \cite{DBLP:conf/sigmod/ZhangRL96}.
The creation of CF-tree hierarchically clusters our data and all the
original data points reside in the leaf nodes of the cluster. These
are also referred to as the micro-clusters \cite{handatamining2nded}. 

In the second phase of the algorithm, which is the novelty of our
approach, we use the multivariate Gaussian density distribution function
\ref{eq:mvnpdf} to split the clusters. There are related research
where the authors use the number of data points \cite{DBLP:conf/sigmod/ZhangRL96}
and variance \cite{DBLP:conf/pakdd/JinGQ06} in a cluster as measures
to split a CF-Tree based cluster. Our rationale behind using the Gaussian
density distribution function is, it will help us better in identifying
those data points whose probability of belonging to a specific micro-cluster
is low. 

The Gaussian density function $f_{X}(x)$ varies from cluster-to-cluster.
We normalize the $f_{X}(x)$ using the min-max normalization. Next
we use a global threshold $\rho$ to split the clusters. The data
values whose $f_{X}(x)$ are below the threshold $\rho$ forms a new
micro-cluster.

\subsection{Details}
\textbf{theorem}
From \cite{DBLP:conf/sigmod/ZhangRL96} - Given $n$
$d$-dimensional data points in a cluster: \{$C_{i}$\} where i =
1, 2, . . . . n, the clustering feature (CF) vector of the cluster
is defined as a triple, \ref{eq:CFeq}. 

\textbf{theorem}
From \cite{DBLP:conf/sigmod/ZhangRL96} \textendash{} Assume that
$CF_{1}=\langle N_{1},LS_{1,}SS_{1}\rangle$, $CF_{2}=\langle N_{2},LS_{2,}SS_{2}\rangle$
are the CF vectors of two disjoint clusters. The CF vector of the
cluster that is formed by merging the two disjoint clusters\textup{
$CF_{1}$} and $CF_{2}$, is: $CF_{1}+CF_{2}=\langle N_{1}+N_{2},LS_{1}+LS_{2},SS_{1}+SS_{2}\rangle$.

\subsubsection{Creation of CF-Tree}
\begin{itemize}
\item The algorithm builds a dendrogram called clustering feature tree (CF-Tree)
while scanning the data set \cite{DBLP:conf/sigmod/ZhangRL96}.
\item Each non-leaf node contains a number of child nodes. The number of
children that a non-leaf node can contain is limited by a threshold
called the branching factor.
\item The diameter of a sub cluster under a leaf node cannot exceed a user
specified threshold.
\item The linear sum and squared sum for a node is already defined in \ref{sub:CF-Tree}.
\item For a non-leaf node, which has child nodes $n_{1},n_{2},\ldots,n_{k}$
\begin{equation}
\vec{{LS}}=\sum_{i=1}^{k}\vec{{LS}}N_{i}\label{eq:LSvec}
\end{equation}
\begin{equation}
\vec{{SS}}=\sum_{i=1}^{k}\vec{{SS}}N_{i}\label{eq:LSvec-1}
\end{equation}

\end{itemize}

\subsubsection{Algorithm}

\paragraph{Phase 1: Building the CF Tree}

\begin{figure}[h!] 
\begin{centering} \scriptsize
   \includegraphics[width=0.45\textwidth]{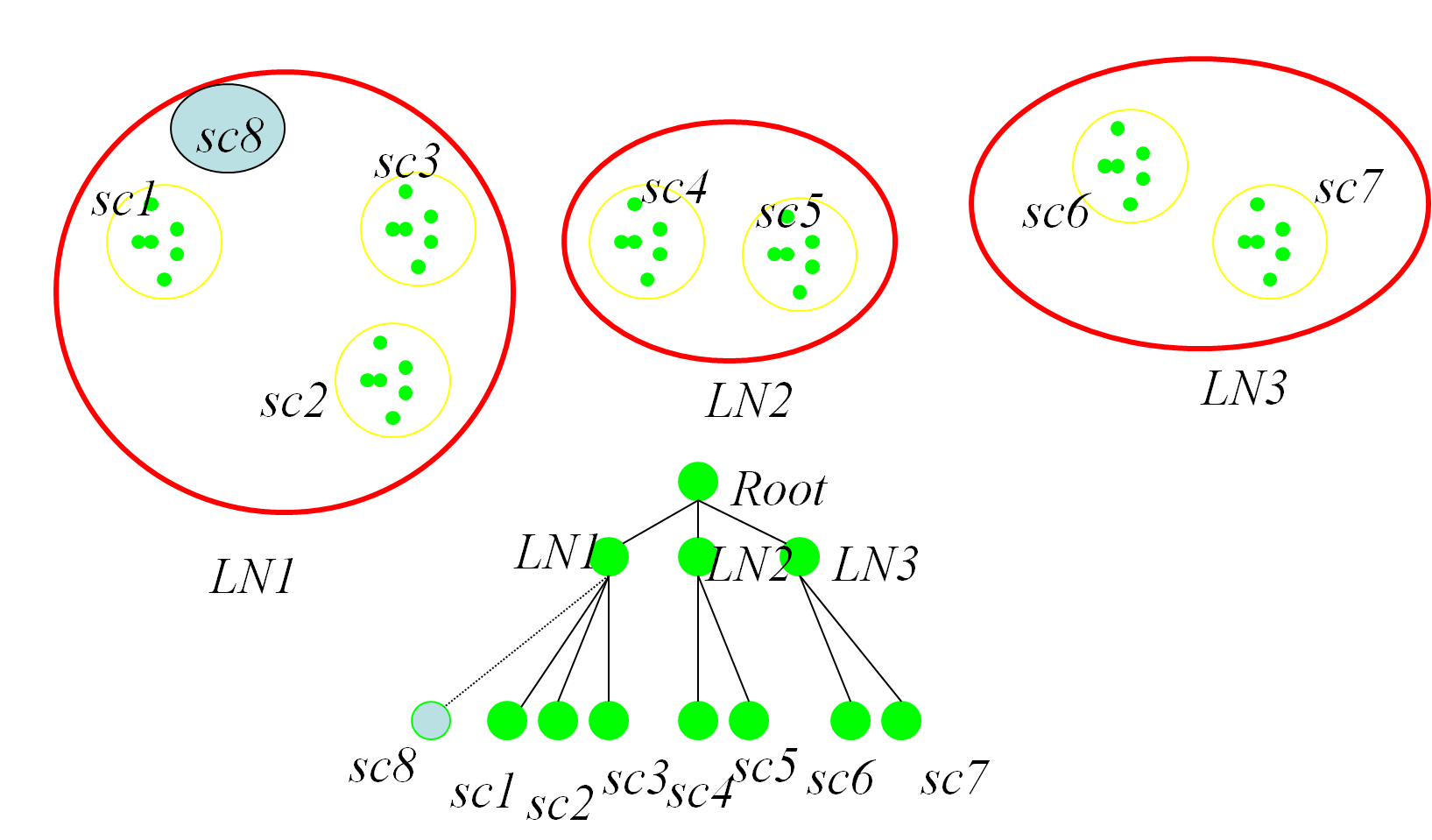}
  \caption{Insertion of a new sub cluster in the CF tree \label{fig:Insertion-of-a}}
    \end{centering}
\end{figure}

The algorithm first scans the data set and inserts the incoming data
instances into the CF tree one by one. The insertion of a data instance
into the CF tree is carried out by traversing the tree top-down from
the root according to an instance-cluster distance function. The data
instance is inserted into the closest sub cluster under a leaf node.
If the insertion of a data instance into a sub cluster causes the
diameter of the sub cluster to exceed the threshold, a new sub cluster
is created and is demonstrated in \ref{fig:Insertion-of-a}.
In this figure the red and the yellow circles denote the clusters. 


\begin{figure}[h!] 
\begin{centering} \scriptsize
   \includegraphics[width=0.45\textwidth]{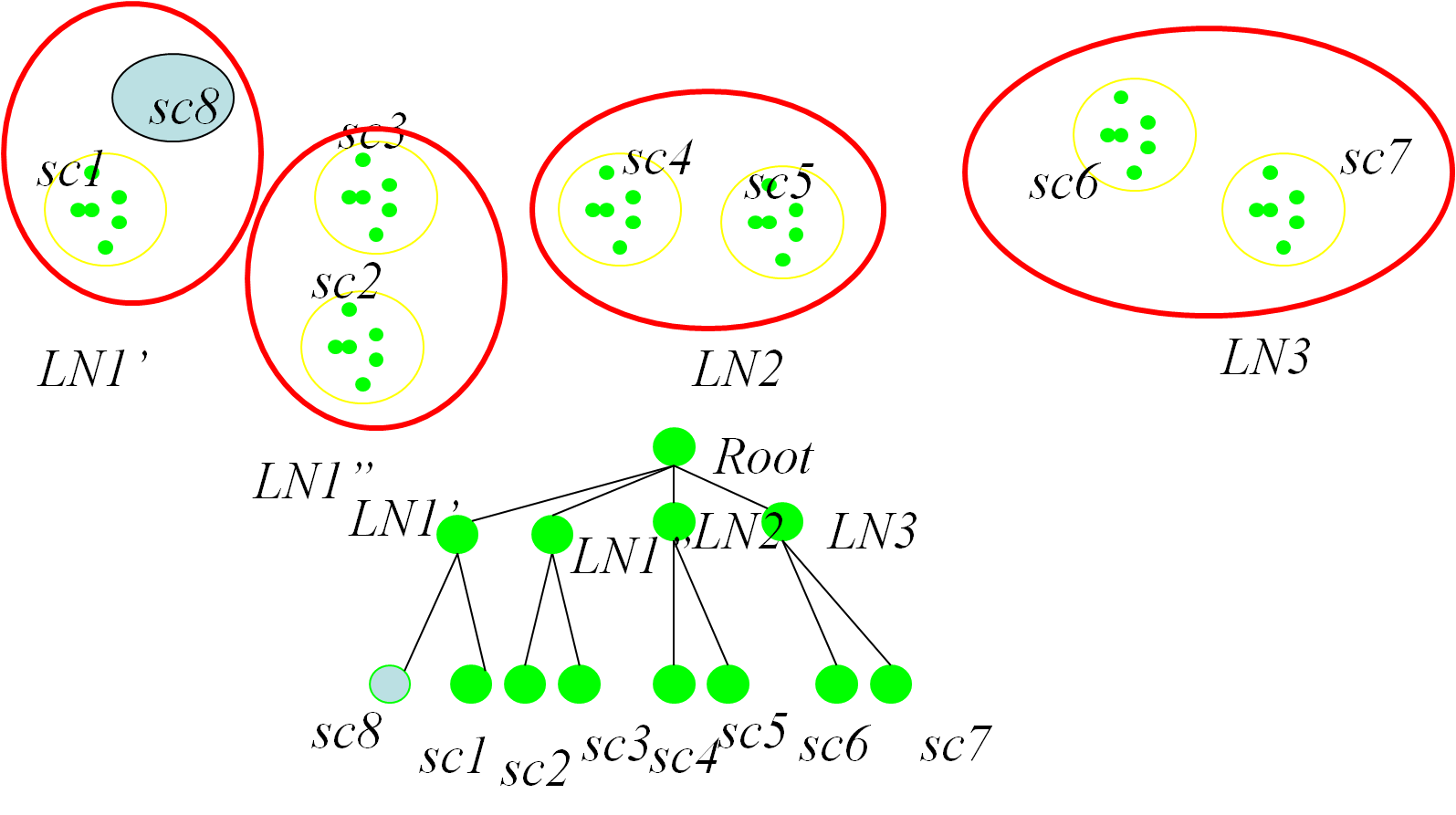}
  \caption{Split of a leaf node if it exceeds branching factor threshold \label{fig:Split-of-a}}
    \end{centering}
\end{figure}

\begin{figure}[h!] 
\begin{centering} \scriptsize
   \includegraphics[width=0.45\textwidth]{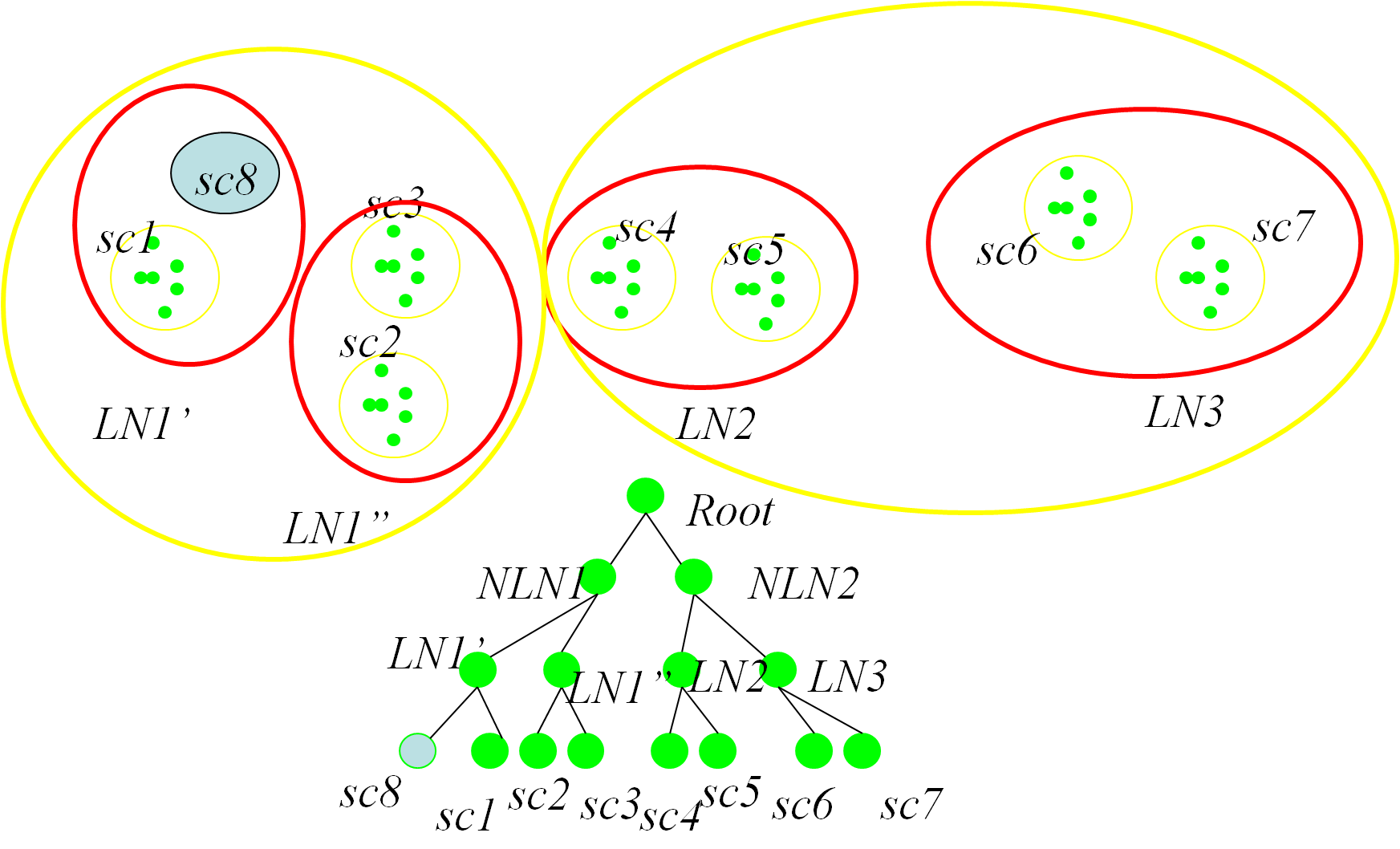}
  \caption{Split of the non-leaf nodes recursively based on inter-cluster distance measure \label{fig:Split-of-the}}
    \end{centering}
\end{figure}

The creation of a new leaf level sub cluster may cause its parent
to cross the branching factor threshold. This causes a split in the
parent node. The split of the parent node is conducted by first identifying
the pair of sub clusters under the node that are separated by the
largest inter-cluster distance. Then, all other sub clusters are dispatched
to the two new leaf nodes based on their proximity to these two sub
clusters. This is demonstrated in \ref{fig:Split-of-a}. Split
of a leaf node may results in a non-leaf node containing more children
than the pre-defined branching factor threshold. If so, the non-leaf
nodes are split recursively based on a measure of inter-cluster distance.
If the root node is split, then the height of the CF tree is increased
by one, as in \ref{fig:Split-of-the}. 


When the split of nodes terminates at a node, merge of the closest
pair of child nodes is conducted as long as these two nodes were not
formed by the latest split. A merge operation may lead to an immediate
split, if the node formed by the merge contains too many child nodes.
Having constructed the CF tree, the algorithm then prepares the CF
tree for the phase 2 of the algorithm to cluster the micro-clusters.

%

\paragraph{Phase 2: Refining using Gaussian constraint measure}

The CF tree obtained in phase1 now consists of clusters with sub clusters
under each leaf node. Now the Gaussian constraint can be applied for
the post processing of the leaf nodes of the CF tree. 

We consider as a constraint the Gaussian density measure which is
the threshold value used to decide the possibility of a data point
in each leaf node of the CF tree to belong to that node, or in other
words in that micro-cluster. 

At first the Gaussian distribution function is calculated for each
data point in each cluster. For this, we need two parametric inputs: 
\begin{enumerate}
\item \textit{Mean of all data point along each attribute}. It is represented
as: $\mu_{i}=\sum_{i=1}^{n}x_{i}/n$, where $n$ is the number of
data points for each attribute in each individual clusters and $x_{i}$
represents the value of the $i^{th}$ attribute of the $x^{th}$ data
point.
\item \textit{Covariance Matrix} is the covariance between every attribute
with all other attributes in an individual clusters. Let $x=\{x_{1},x_{2},\ldots,x_{p}\}$
be a random vector with a mean of $\mu=\{\mu_{1},\mu_{2},\ldots,\mu_{p}\}$.
\end{enumerate}
\begin{equation}
\Sigma=\left(\begin{array}{cccc}
\sigma_{11} & \sigma_{12} & \ldots & \sigma_{1p}\\
\sigma_{21} & \sigma_{22} & \ldots & \sigma_{2p}\\
\vdots & \vdots & \ddots & \vdots\\
\sigma_{p1} & \sigma_{p2} & \ldots & \sigma_{pp}
\end{array}\right)\label{eq:covmatrix}
\end{equation}

\begin{figure}[h!] 
\begin{centering} \scriptsize
   \includegraphics[width=0.45\textwidth]{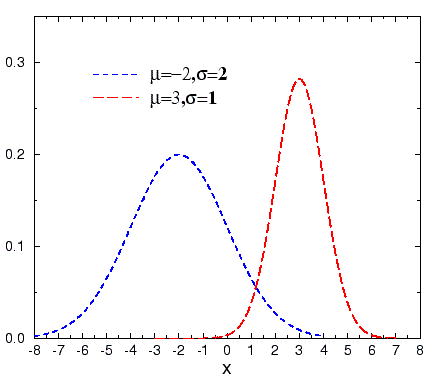}
  \caption{Univariate Gaussian distribution \label{fig:Univariate-Gaussian-distribution}}
    \end{centering}
\end{figure}

The co-variance matrix is denoted as $\Sigma$. $\sigma_{ij}$ is
the co-variance of the $i^{th}$attribute with $j^{th}$attribute.
Thus $\sigma_{ii}$ is the co-variance of the $i^{th}$ attribute
with itself which are all the diagonal elements in \ref{eq:covmatrix}.
An univariate Gaussian distribution is shown in \ref{fig:Univariate-Gaussian-distribution}.

%

The next step is to compute the Gaussian distribution function for
each data point in each cluster using mean and covariance matrix calculated
in the previous step. 

A threshold is decided which iteratively checks all data points in
each cluster to see if they fall within the threshold or not. If some
data point is found to be far away from the threshold, the particular
cluster to which it belongs to currently, is split. In this way all
the data points are compared with the threshold. Two new clusters
are formed with group of data points which have a different range
of Gaussian density measure. Hence, there is possibility of forming
new sets of clusters from the leaf node of the CF tree thereby increasing
its depth. The clusters formed in this way are hypothesized to be
more accurate than that formed in phase 1, since we are comparing
the Gaussian density measure of each data points for each cluster.

\section{Experimental Results}

\subsection{Data Set}

In order to evaluate the performance of our algorithm, we conducted
a number of experiments using Abalone data set published in the year
1995 \cite{Frank+Asuncion:2010:UCI}. Out of the eight dimensions
in the original database, we chose seven continuous dimensions (\textit{Length},
\emph{Diameter}, \emph{Height}, \emph{Whole weight}, \emph{Shucked weight},
\emph{Visceral weight}, \emph{Shell weight}). The eighth dimension,
\emph{Sex}, is a nominal dimension and can take only 3 distinct values.
Thus we excluded it from our experiments. In our data set we have
4177 instances each representing a single abalone inside a whole colony.
The primary goal of the data set is to predict the age of an abalone
from its physical measurements. Although this data set is primarily
used in classification, we use it as supervised evaluation technique
for cluster validity \cite{tan2005introduction}.

\subsection{Micro-Cluster Analysis}

\begin{figure}[h!] 
\begin{centering} \scriptsize
   \includegraphics[width=0.45\textwidth]{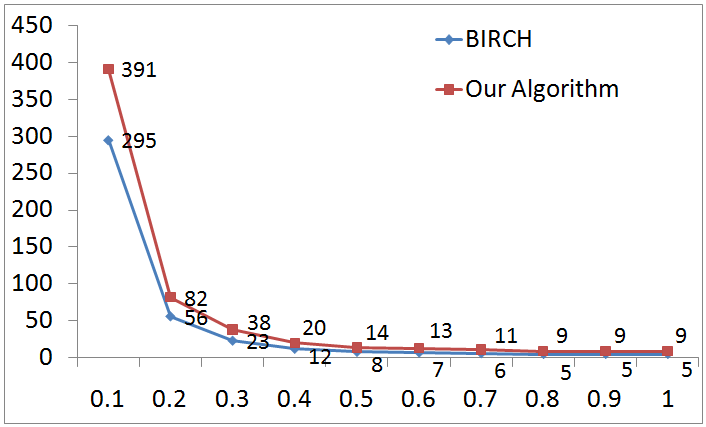}
  \caption{Micro-Cluster Analysis \label{fig:Micro-Cluster-Analysis}}
    \end{centering}
\end{figure}

We present the results of our micro-cluster analysis in \ref{fig:Micro-Cluster-Analysis}.
For the purposes of this test we created the initial CF-tree with
varied distance measure. We started our test with a distance measure
of 0.1 and went all the way to 1.0 with an increase of 0.1 in each
step. The primary purpose of this test is to hypothesize on the trend
of number of micro-clusters our algorithm will generate for a specific
CF-tree input. The X-axis in \ref{fig:Micro-Cluster-Analysis}
represents the distance measure and the Y-axis represents the number
of micro-clusters generated. In this test we compare the number of
micro-clusters generated by BIRCH \cite{DBLP:conf/sigmod/ZhangRL96}
with our approach. 

%

From \ref{fig:Micro-Cluster-Analysis} we can infer that, when
the number of micro-clusters generated by BIRCH is low, our algorithm
generates almost twice the number of micro-clusters. But when the
number of micro-clusters generated by BIRCH increases, the ratio between
the number of micro-clusters generated by our algorithm and BIRCH
tends to 1. The outcome of this analysis is expected, since as the
number of micro-clusters generated by BIRCH increases, the amount
of data points inside the micro-clusters tend to decrease significantly.
When BIRCH generates a relatively large number of micro-clusters,
most of the micro-clusters contain a very few data points in them
which does not cross the threshold we have set in our algorithm for
splitting a micro-cluster.

\subsection{Scalability Test}

\begin{figure}[h!] 
\begin{centering} \scriptsize
   \includegraphics[width=0.50\textwidth]{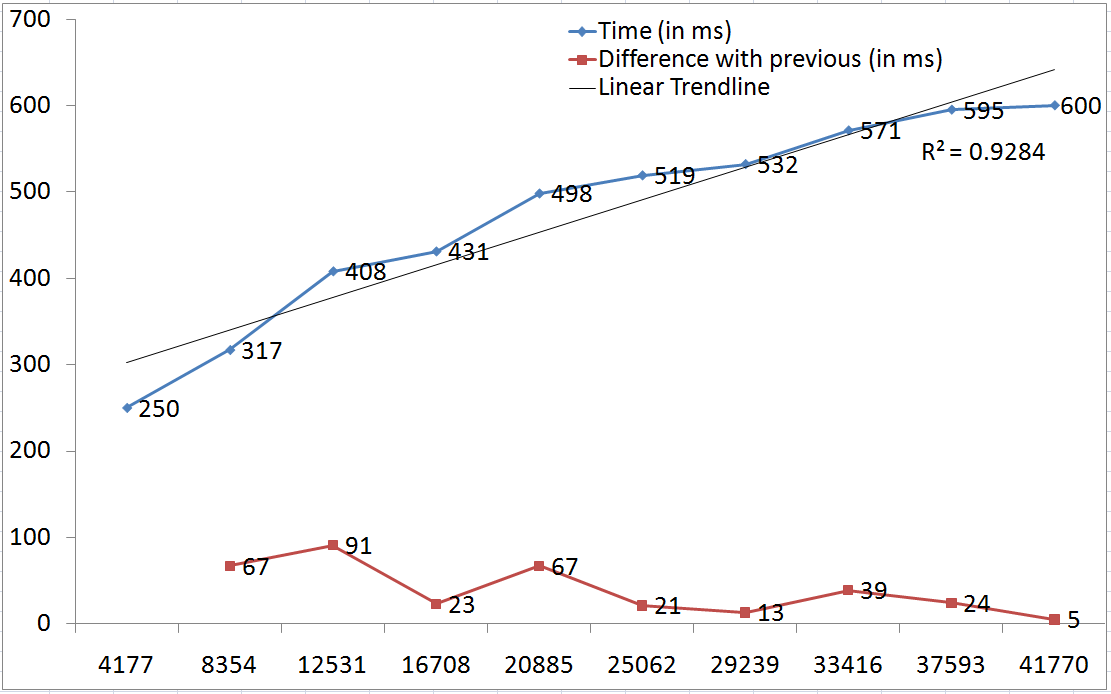}
  \caption{Scalability Test \label{fig:Scalability-Test}}
    \end{centering}
\end{figure}

For the scalability test we use the same data set. Since the number
of tuples in the data set is not elephantine, we scale our data by
appending an instance of the data set at the end of itself. In \ref{fig:Scalability-Test}-
X-axis denotes the number of records in our data set, whereas Y-axis
represents its corresponding running times. In this way we can generate
data sets which are multiples of the original data set. We present
our results in figure \ref{fig:Scalability-Test}. 

%

From the results we can infer that our algorithm has a constant time
component and a linearly increasing component. To lend credibility
to our inference we have also provided the difference in time between
two consecutive results. This curve follows a horizontal straight
line meaning the difference is constant.

\subsection{Supervised Measures of Cluster Validity}

To test the accuracy of our model we have compared our algorithm with
two related approaches: BIRCH \cite{DBLP:conf/sigmod/ZhangRL96} and
CDC \cite{DBLP:conf/kdd/GeEJD07}. We have compared our results based
on the measures presented in \cite{tan2005introduction} namely entropy,
purity, precision and recall. For this analysis we set the distance
measure in BIRCH to 0.27 so that it creates 30 micro-clusters. In
response to BIRCH, our algorithm created 47 micro-clusters. We tuned
the experiment in a such a way that BIRCH would create almost the
same number of clusters as the number of class labels in the original
data set. This provided BIRCH with a slight yet distinct advantages
over our approach. The CDC algorithm was also tuned to create approximately
the same number of clusters as the original class labels. To achieve
this we set the significance measure in CDC to 80 and a variance of
0.34. A significance measure of 80 ensures that CDC will have between
80 to 179 nodes in all its micro-clusters. The summarized results
of entropy and purity for the 3 approaches are presented in \ref{tab:Supervised-Measures}.
\begin{table}[tbh]
\protect\caption{Supervised Measures \label{tab:Supervised-Measures}}

\centering{}%
\begin{tabular}{|c|c|c|}
\hline 
 & Entropy & Purity\tabularnewline
\hline 
\hline 
BIRCH & 2.415 & 0.389\tabularnewline
\hline 
Our Algorithm & 2.959 & 0.499\tabularnewline
\hline 
CDC & 3.521 & 0.601\tabularnewline
\hline 
\end{tabular}
\end{table}

From the table we can infer that for this specific data set, our algorithm
provides comparatively better results than BIRCH although our algorithm
started the analysis with a slight disadvantage. As expected CDC out
performs both the algorithms in this test. The detailed result table
can be found in \cite{appendix}.

\section{Conclusion and Future Work}

In this paper we present an efficient multiphase iterative hierarchical
constraint based clustering technique. Our algorithm is based on the
clustering feature tree and uses Gaussian probabilistic density based
distribution measure as a constraint for refining the micro-clusters.
It efficiently formed clusters and ended up with a better result than
the traditional BIRCH algorithm for this particular data set. Also
our algorithm gives, comparable results when pitted against the CDC
algorithm \cite{DBLP:conf/kdd/GeEJD07}. The scalability result shows
that our algorithm performs efficiently even when tested with large
data sets. We tested our algorithm and BIRCH with the same data set,
and our algorithm produced more micro-clusters than BIRCH. Yet the
entropy and precision ofour algorithm was higher than that of BIRCH.
This indicates the accuracy of our algorithm.

We have ample scope for future work in this area. Currently our algorithm
undergoes only one iteration in phase 2 for the refinement of the
micro clusters. But we plan to implement iterative refinement procedure
using the Gaussian density distribution constraint, for a better result.
This iteration in phase 2 would continue until no more clusters can
be formed. Also, we need to implement a merging algorithm which would
merge the huge number of micro-clusters we hypothesize our future
algorithm will form. Another important task that we consider as our
future work is to test our algorithm with data sets having non continuous
class level. 


\section*{Acknowledgment}
We would like to thank Prof. Rafal Angryk for his feedback, support and critical reviews during developing of this project.

\bibliographystyle{amsplain}
\bibliography{litreview, scibib}

\end{document}